\documentclass[dvipsnames,format=sigconf,nonacm=true]{acmart}

\usepackage[inline]{enumitem}
\usepackage{pdfpages}
\usepackage{pdflscape}

\begin{document}
\title{Towards Exploratory Quality Diversity Landscape Analysis}

\author{Kyriacos Mosphilis}
\affiliation{%
    \institution{CYENS Centre of Excellence}
    \city{Nicosia}
    \country{Cyprus}}
\email{k.mosphilis@cyens.org.cy}

\author{Vassilis Vassiliades}
\affiliation{%
    \institution{CYENS Centre of Excellence}
    \city{Nicosia}
    \country{Cyprus}}
\email{v.vassiliades@cyens.org.cy}

\begin{abstract}
This work is a preliminary study on using \emph{Exploratory Landscape Analysis} (ELA) for \emph{Quality Diversity} (QD) problems. We seek to understand whether ELA features can potentially be used to characterise QD problems paving the way for automating QD algorithm selection. Our results demonstrate that ELA features are affected by QD optimisation differently than random sampling, and more specifically, by the choice of variation operator, behaviour function, archive size and problem dimensionality.
    % Many recent works have introduced many different operators for \emph{Quality Diversity} (QD). These operators, however, usually outperform others in specific problems, where the elite-hypervolume $E$ of the problems has a characteristic that differs that other problems. As the $E$ is usually unknown, and becomes increasingly difficult to understand as the dimensions increase, perhaps understanding the $E$ could prove to be beneficial. In this paper, we use features introduced by the field of \emph{Exploratory Landscape Analysis}, where these features will help us characterise $E$.
    % a black-box optimisation problem, so that we can understand
    % \begin{enumerate*}
    %     \item the nature of a QD problem, and
    %     \item how this nature can affect the behaviour of the operators.
    % \end{enumerate*}
    % in such way to understand which is the most suitable approach to solve it.
\end{abstract}

\maketitle

\section{Introduction}

Exploratory Landscape Analysis \cite{mersmann2010benchmarking, mersmann2011exploratory} aims at designing features that can characterise properties of an objective function from a few samples. Having these features, as well as knowledge of which algorithms are better suited for which problem group, one may choose the right algorithm to solve the problem efficiently.
Examples of expert-designed properties (or high-level features) of functions are: multi-modality, global structure, separability, variable scaling, search space homogeneity, basin size homogeneity, global to local optima contrast, and plateaus \cite{mersmann2010benchmarking}. These properties (with the exception of the latter) have been used to classify the noiseless BBOB benchmark functions \cite{hansen2010real} into groups. 

However, as these properties are not known for a problem at hand, ELA in its original version proposed 50 low-level features grouped into six classes: convexity, curvature, y-distribution, levelset, local search, and meta model \cite{mersmann2011exploratory}. These low-level feature classes can be used to characterise the high-level properties, and are computed by obtaining a data set $D$ of $m$ samples from the genotype space (decision variables) $\mathbf{x^i} \in \mathbb{R}^d$ with their associated objective function values $y^i=f(\mathbf{x^i})$, i.e., $D = \{\mathbf{x^i},y^i\}_{i=1:m}$, using Latin hypercube sampling (LHS). Additional classes of low-level features have been proposed such as cell mapping \cite{kerschke2014cell}, dispersion \cite{lunacek2006dispersion}, information content-based \cite{munoz2014exploratory}, and nearest-better \cite{kerschke2015detecting} features.

ELA coupled with automated algorithm selection hold promise in accelerating optimization for real-world problems, where often the goal is to obtain a single optimal solution. 
There are certain types of problems, however, where the goal is to obtain a large and diverse archive of high-performing solutions. These problems are the focus of Quality-Diversity (QD) optimization \cite{pugh2016quality, cully2017quality, chatzilygeroudis2021quality}, and examples include generating behavioral repertoires for robots \cite{cully2015robots}, content for video games \cite{gravina2019procedural}, airfoil designs \cite{gaier2017data} and others.
A key component of QD methods is that they maintain diversity in a space different than the genotype space, i.e., the phenotype or behavior space. The interaction of this ``diversity space'' (or behaviour function) with the objective function, has provided evidence that it leads to concentrations of solutions in the genotype space, called the Elite Hypervolume \cite{vassiliades2018discovering}, which could potentially be exploited to accelerate QD optimization \cite{vassiliades2018discovering, fontaine2020covariance, christou2023quality}.

The central question behind this work is whether we could use ELA to characterise QD problems, thus, pave the way for automating algorithm (or operator) selection in QD optimisation. Our main hypothesis is that QD optimisation affects the low-level ELA features differently than LHS. Therefore, we set out to investigate how these features are affected by: (1) the choice of variation operator over the course of generations, (2) the choice of behaviour function for the same problem, (3) different archive sizes, and (4) different problem dimensionalities.

% We didn't include the bounding space for the Robot Arm. Should we included it ([-pi,pi])?
\section{Experiments}
We performed $30$ experimental runs using $3$ domains (\emph{Robot Arm Repertoire} \cite{vassiliades2018discovering}, the \emph{Sphere} and the \emph{Rastrigin} functions), $3$ algorithms for ``sampling solutions'' (LHS, CVT-MAP-Elites \cite{vassiliades2017using} with either Gaussian mutations or the IsoLineDD operator \cite{vassiliades2018discovering}), $5$ different genotypic dimensions ($\{2, 4, 8, 16, 32\}$), phenotypic dimensions ($[-12,12]$ for \emph{Robot Arm}, and $[-5,5]$ for the other two), and $3$ archive sizes ($\{100, 1000, 10000\}$). In addition, for the \emph{Sphere} and \emph{Rastrigin} functions we chose 3 different behaviour functions as $\mathbf{b^i}=\phi(\mathbf{x^i}) \in \mathbb{R}^2$:
\begin{enumerate*}
    \item $\mathbf{b^i}=[x_0, x_1]^T$ (subset of the first two dimensions), normalised in the range $[0,1]^2$,
    \item $\mathbf{b^i}=$ $sig(\mathbf{x^T} W)^T$, where $W \in \mathbb{R}^{d \times 2}$ is a randomly initialised weight matrix, and $sig$ is the sigmoid function,
    \item $\mathbf{b^i}=sine(\mathbf{x^T} W)^T$, where $sine$ is the sine function, normalised in the range $[0,1]^2$.
    % \item two Artificial Neural Networks (ANNs); one for $x$ and one for $y$ with only one layer --- the output $x$ and $y$ respectively, randomly initialised weights, and the \emph{Sigmoid} activation function, and
    % \item two ANNs similar to the aforementioned, with the only difference being the activation function, as we chose the \emph{Sine} function to provide the output.
\end{enumerate*}
For the CVT-MAP-Elites variants, we have performed $1\mathrm{e}+6$ evaluations --- $1\mathrm{e}+4$ generations with $100$ evaluations per generation.
We use the pyribs library~\cite{pyribs} for the QD algorithms and pflacco~\cite{pflacco} for the ELA features (selected calculated features at~\autoref{table:features}).

% With these experiments we aim to analyse and understand \emph{if} and \emph{how} the ELA features are affected with the change of the \emph{\textbf{dimensions}}, the \emph{\textbf{total samples (archive size)}}, the usage of a different \emph{\textbf{behaviour function}}, and when a different \emph{\textbf{operator}} is used for new solutions.

% Change natures of the feature. and also the samples sizes to also dimensions. Be more specific.
In the following results and discussions, some ELA features could not be calculated due to a lack of samples, some others required a large amount of time to be calculated as the size of the samples and the archive were increasing (mainly meta-model~\cite{mersmann2011exploratory}), some are affected by dimensionality (cell mapping), and some features that calculate gradients and Hessian matrices (curvature~\cite{mersmann2011exploratory}, and cell mapping) require a significant amount of time, even when using the smallest sample/archive size.

\subsection{Effect of variation operator}
\begin{landscape}
    \begin{figure}[ht]
        \centering
        \includegraphics[scale=0.74]{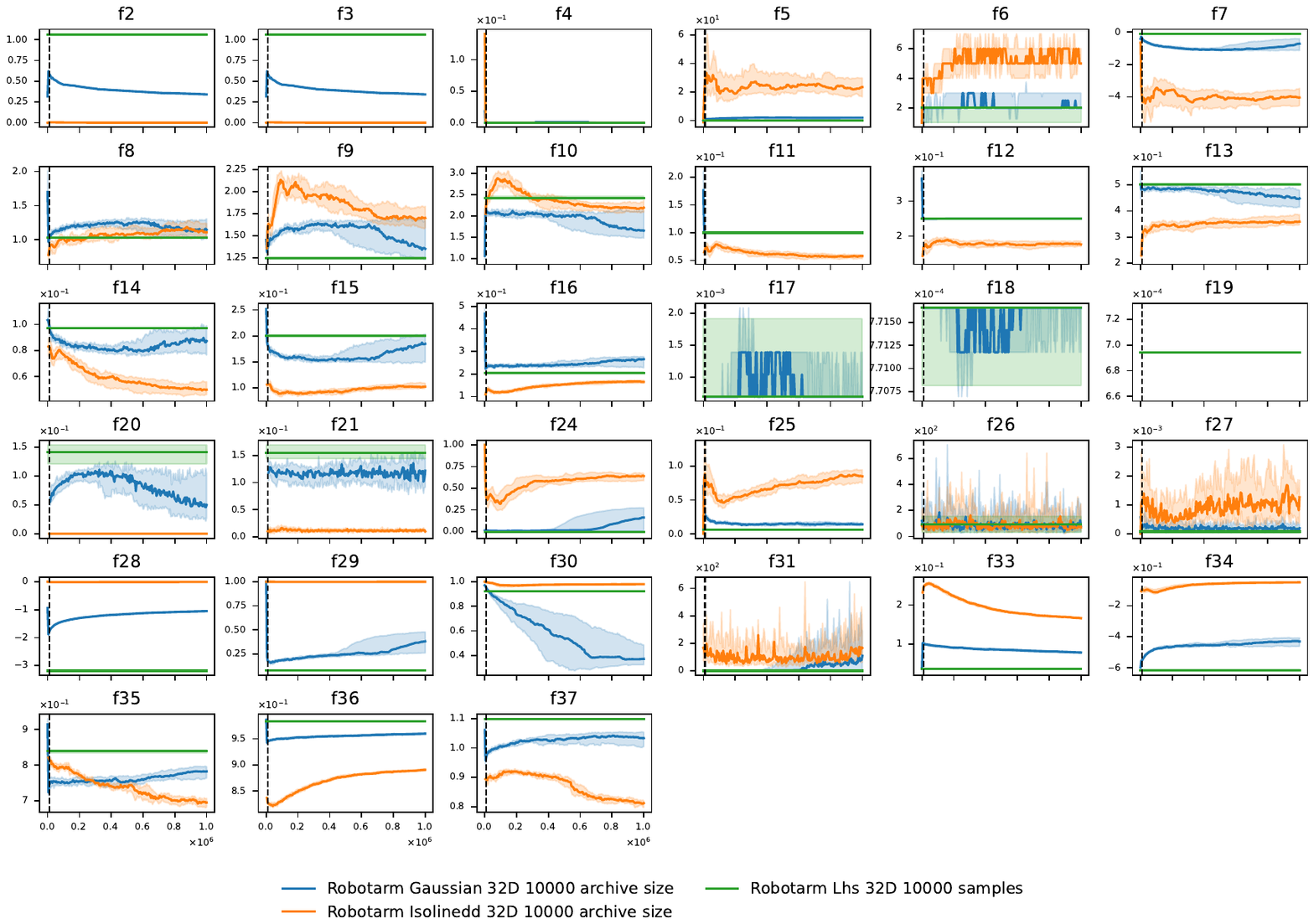}
        \caption{Gaussian mutation operator and IsoLineDD in comparison with LHS with $10000$ archive cells on the \emph{Robot Arm Repertoire} domain with $32$ dimensions. The plots show the median and the interquartile range ($Q_1$ and $Q_3$) over $30$ experimental runs, with mostly notable differences. The x-axis, represents the total number of evaluations, whilst the vertical dashed line marks the $10000$ evaluations, matching the $10000$ evaluations of LHS.}
        \Description{Gaussian multivariate operator and IsoLineDD in comparison with LHS with $10000$ archive cells on the \emph{Robot Arm Repertoire} domain with $32$ dimensions. The plots show the median and the interquartile range ($Q_1$ and $Q_3$) over $30$ experimental runs, with mostly notable differences. The x-axis, represents the total number of evaluations, whilst the vertical dashed line marks the $10000$ evaluations, matching the $10000$ evaluations of LHS.}
        \label{fig:appendix:operator:dim32}
    \end{figure}
\end{landscape}

We compared the Gaussian mutation operator with the IsoLineDD operator while using the LHS method as a baseline. We have included two plots in~\autoref{appendix:figures}. Both are for archive size 10000 (up to 10000 samples), one for 16 dimensions (~\autoref{fig:appendix:operator:dim16}) and one for 32 dimensions (\autoref{fig:appendix:operator:dim32}). By inspecting the plots and the results, we have noticed that some of the results demonstrate a significant difference (\autoref{fig:appendix:operator:dim16}, f5: $p$-value $< 1\mathrm{e}-10$ for all pairs, Mann-Whitney U test), and some that are identical 
(\autoref{fig:appendix:operator:dim32}, f19: $p = 1$ for all pairs, Mann-Whitney U test).
% Moreover, features like f8 - f10 and f14 - f16, show that features are susceptible to the change of the archive over generations (and thus evaluations). 

\subsection{Effect of behaviour function}
\begin{landscape}
    \begin{figure}[ht]
        \centering
        \includegraphics[scale=0.74]{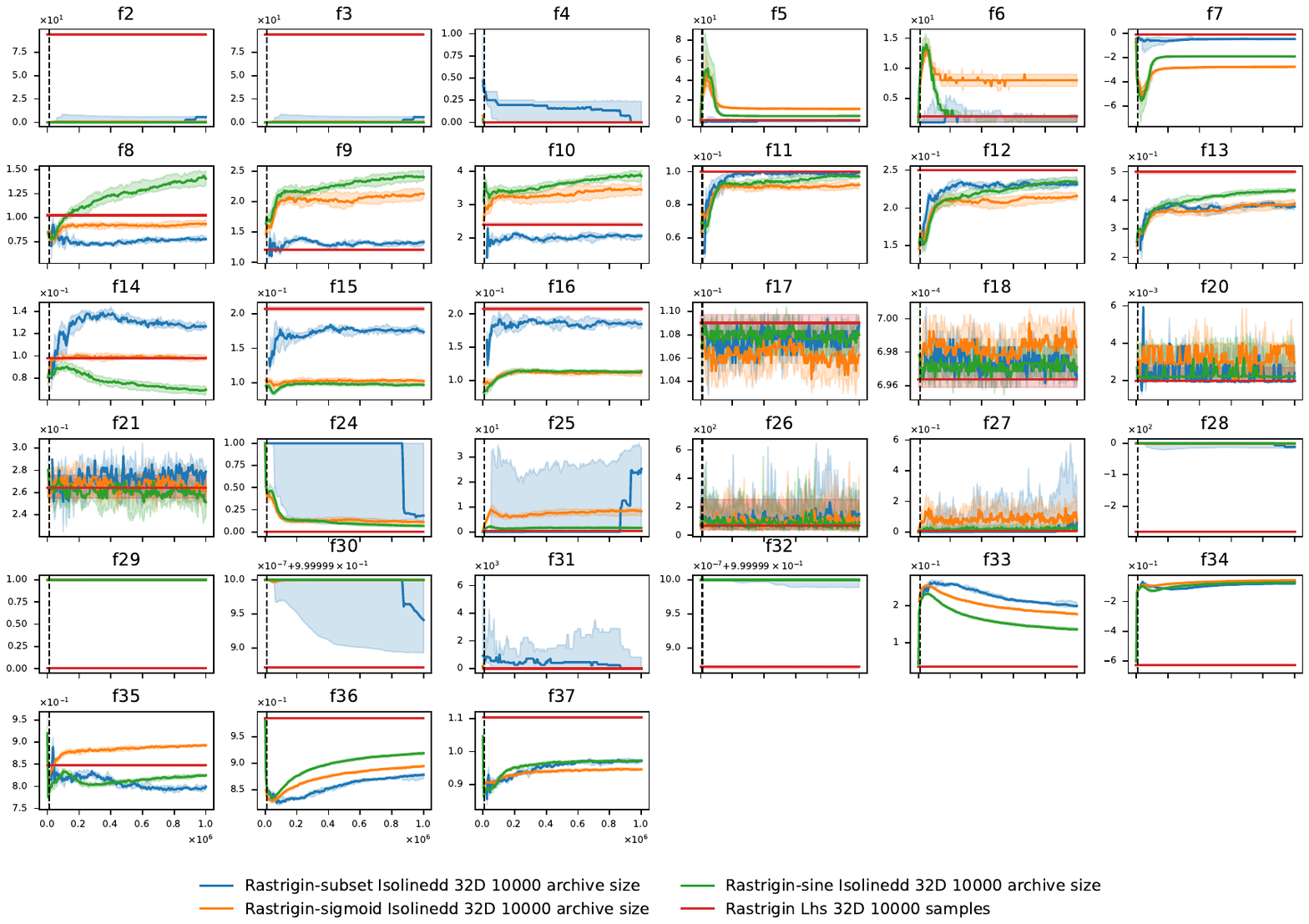}
        \caption{IsoLineDD in comparison with LHS on the \emph{Rastrigin} function with all the different behaviour functions with $10000$ archive size and $32$ dimensions. The plots show the median and the interquartile range ($Q_1$ and $Q_3$) over $30$ experimental runs, with mostly notable differences. The x-axis, represents the total number of evaluations, whilst the vertical dashed line marks the $10000$ evaluations, matching the $10000$ evaluations of LHS.}
        \Description{IsoLineDD in comparison with LHS on the \emph{Rastrigin} function with all the different behaviour functions with $10000$ archive size and $32$ dimensions. The plots show the median and the interquartile range ($Q_1$ and $Q_3$) over $30$ experimental runs, with mostly notable differences. The x-axis, represents the total number of evaluations, whilst the vertical dashed line marks the $10000$ evaluations, matching the $10000$ evaluations of LHS.}
        \label{fig:appendix:behaviour:rastrigin}
    \end{figure}
\end{landscape}

\begin{landscape}
    \begin{figure}[ht]
        \centering
        \includegraphics[scale=0.74]{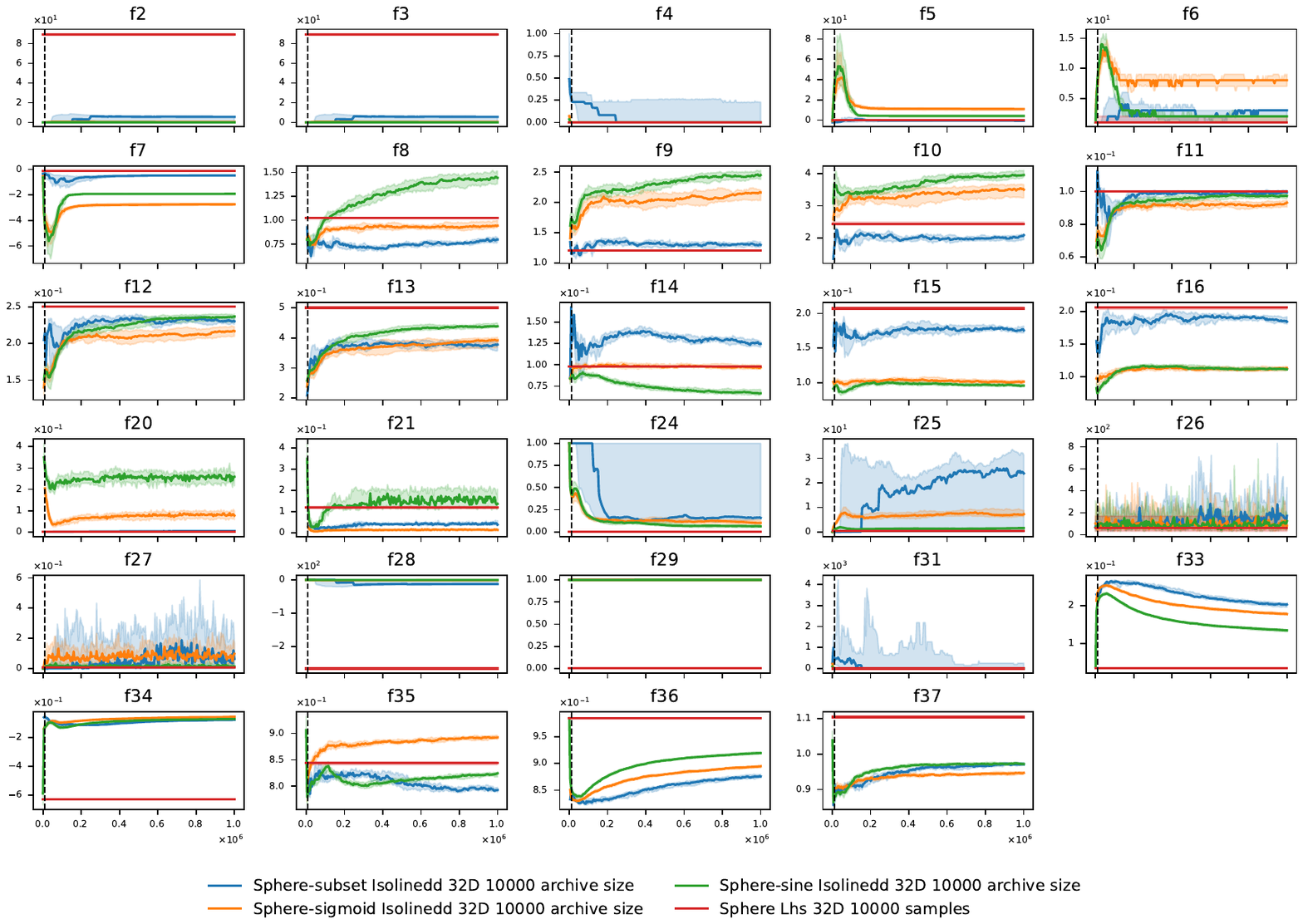}
        \caption{IsoLineDD in comparison with LHS on the \emph{Sphere} function with all the different behaviour functions $\mathbf{b^i}$ with $10000$ archive size and $32$ dimensions. The plots show the median and the interquartile range ($Q_1$ and $Q_3$) over $30$ experimental runs, with mostly notable differences. The x-axis, represents the total number of evaluations, whilst the vertical dashed line marks the $10000$ evaluations, matching the $10000$ evaluations of LHS.}
        \Description{IsoLineDD in comparison with LHS on the \emph{Sphere} function with all the different behaviour functions $\mathbf{b^i}$ with $10000$ archive size and $32$ dimensions. The plots show the median and the interquartile range ($Q_1$ and $Q_3$) over $30$ experimental runs, with mostly notable differences. The x-axis, represents the total number of evaluations, whilst the vertical dashed line marks the $10000$ evaluations, matching the $10000$ evaluations of LHS.}
        \label{fig:appendix:behaviour:sphere}
    \end{figure}
\end{landscape}

% Maybe show f15, f16, or f34 with sine and sigmoid
For the \emph{Sphere} and the \emph{Rastrigin} functions, we have chosen 3 different behaviour functions. We compared these behaviour functions using the IsoLineDD operator with 10000 archive size, and 32 dimensions; \autoref{fig:appendix:behaviour:sphere} --- \emph{Sphere}, \autoref{fig:appendix:behaviour:rastrigin} --- \emph{Rastrigin}.
% We have included the effect of different $\mathbf{b^i}$ plots of \emph{Sphere}~\autoref{fig:appendix:behaviour:sphere} and \emph{Rastrigin}~\autoref{fig:appendix:behaviour:rastrigin} on IsoLineDD with 10000 archive size and 32 dimensions in the appendix. 
Whilst most of them change over time, certain features display fluctuation over time (f14 - f16 in both figures). We have also noticed some significant differences (\autoref{fig:appendix:behaviour:rastrigin} f10: $sig$ with $sine$, $p < 1\mathrm{e}-6$ Mann-Whitney U test), but also some similarity (\autoref{fig:appendix:behaviour:rastrigin} f16: $sig$ with $sine$, $p = 0.41$, Mann-Whitney U test).

\subsection{Effect of archive size}
\begin{figure*}[ht]
    \centering
    \includegraphics[scale=0.6745]{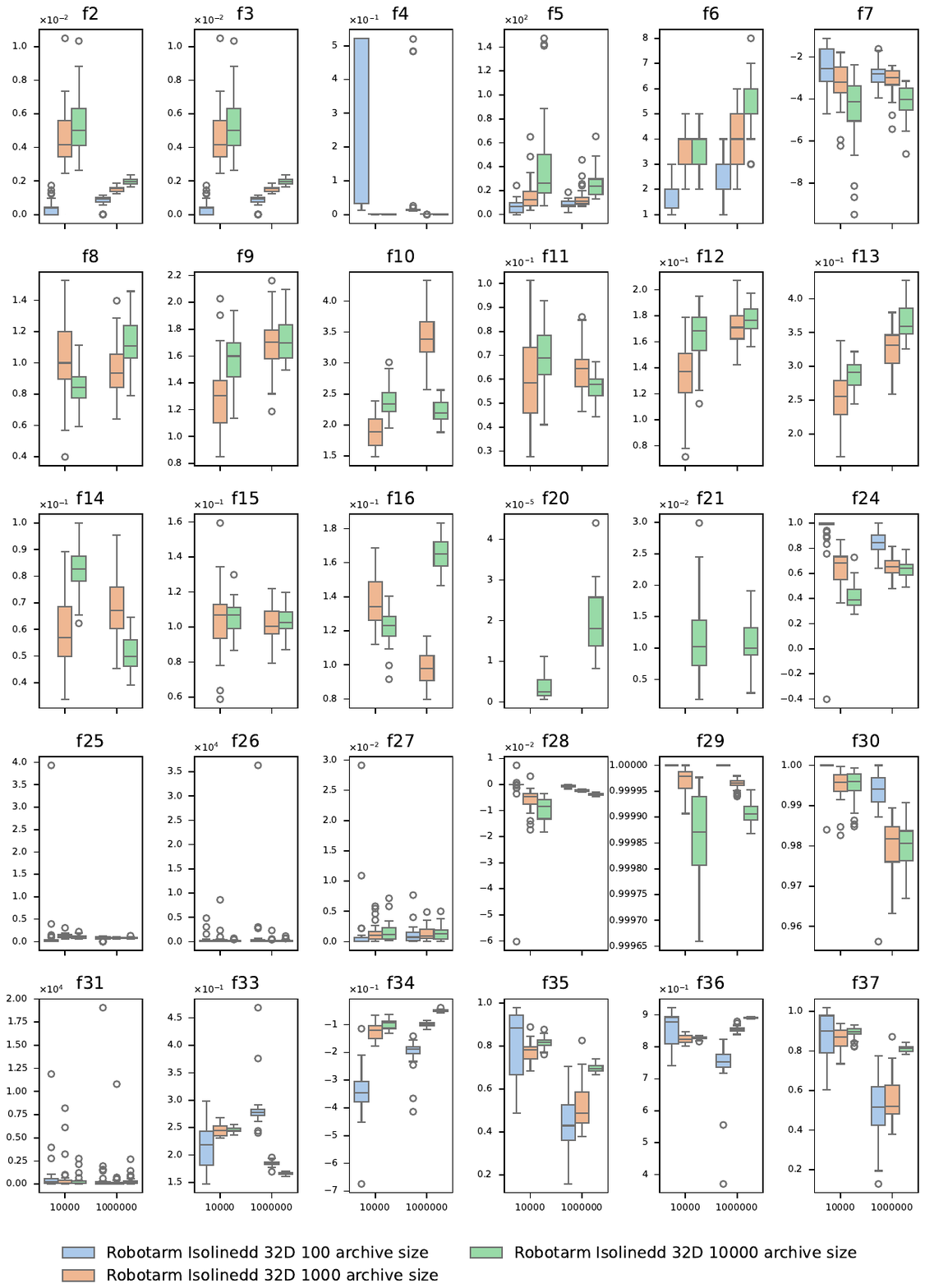}
    \caption{IsoLineDD on the \emph{Robot Arm Repertoire} with $32$ dimensions on different archive sizes. The plots show the state of the features at $10k$ evaluations on different archive sizes (notice the different colours), and how it changes at $1M$ evaluations. The box plots are groups based on theses two evaluation milestones. Some features could not be calculated for $100$ archive size; low sample count.}
    \Description{IsoLineDD on the \emph{Robot Arm Repertoire} with $32$ dimensions on different archive sizes. The plots show the state of the features at $10k$ evaluations on different archive sizes (notice the different colours), and how it changes at $1M$ evaluations. The box plots are groups based on theses two evaluation milestones. Some features could not be calculated for $100$ archive size; low sample count.}
    \label{fig:appendix:size:robotarm}
\end{figure*}

By comparing the IsoLineDD algorithm with 32 dimensions on the \emph{Robot Arm Repertoire} using all archive sizes (\autoref{fig:appendix:size:robotarm}), we noticed some significant differences on many features (\autoref{fig:appendix:size:robotarm} f29: $p < 1\mathrm{e}-15$ in all pairs, Mann-Whitney U test). We have also noticed that the features extracted after $1M$ evaluations, have a significantly different state, than after $10k$ evaluations (\autoref{fig:appendix:size:robotarm} f33: 10000 archive size: $10k$ with $1M$ $p < 1\mathrm{e}-10$, Mann-Whitney U test), but interestingly enough, we have encountered features that do not show significant change (\autoref{fig:appendix:size:robotarm} f27: 10000 archive size: $10k$ with $1M$ $p < 0.43$, Mann-Whitney U test).

\subsection{Effect of problem dimension}
\begin{figure*}[ht]
    \centering
    \includegraphics[scale=0.6875]{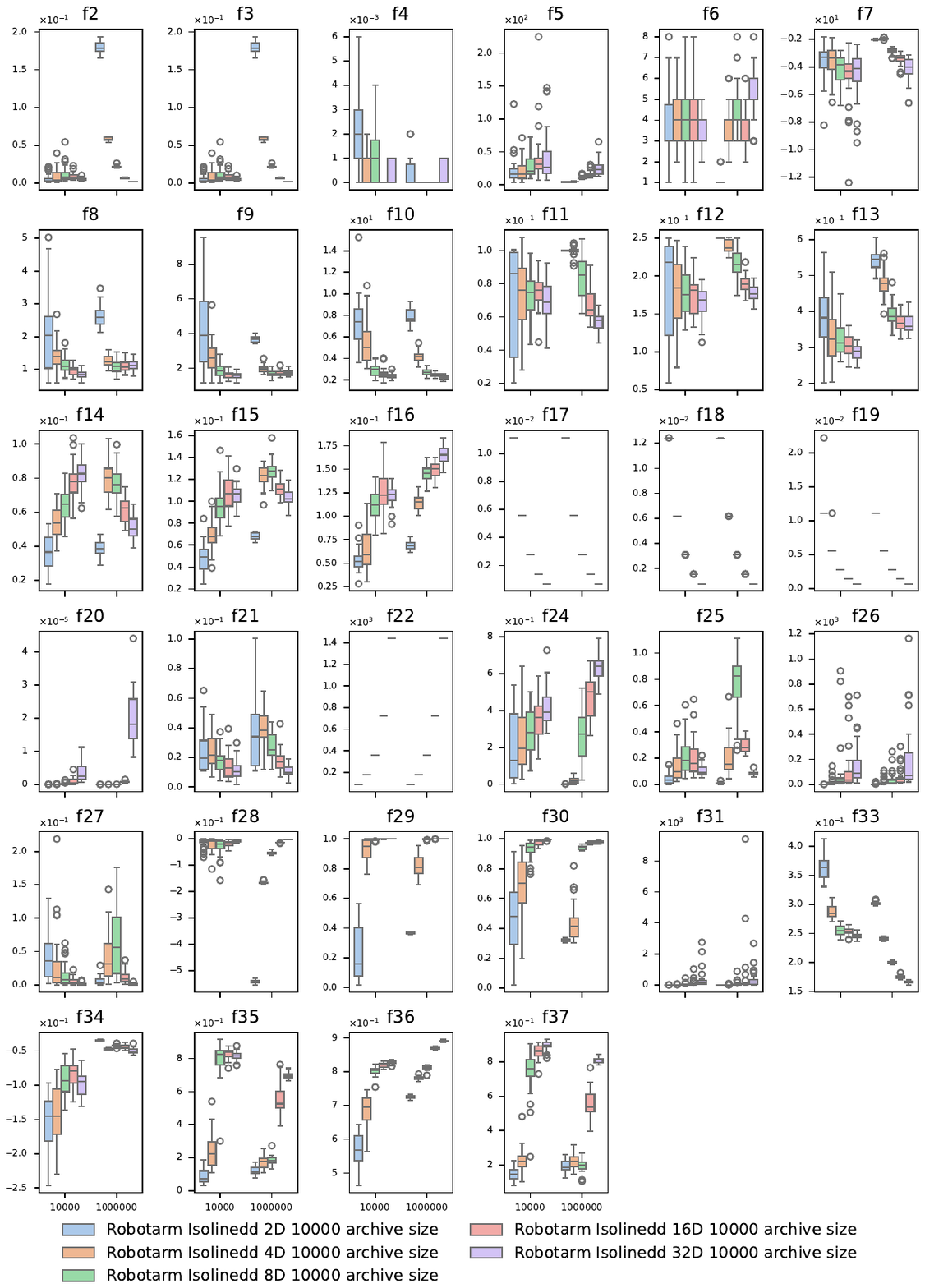}
    \caption{IsoLineDD on the \emph{Robot Arm Repertoire} with $10000$ archive size on different dimensions. The plots show the state of the features on $10k$ evaluations on different dimensions (notice the different colours), and how it changes on the $1M$ evaluations. The box plots are groups based on theses two evaluation milestones.}
    \Description{IsoLineDD on the \emph{Robot Arm Repertoire} with $10000$ archive size on different dimensions. The plots show the state of the features on $10k$ evaluations on different dimensions (notice the different colours), and how it changes on the $1M$ evaluations. The box plots are groups based on theses two evaluation milestones.}
    \label{fig:appendix:dim:robotarm}
\end{figure*}

Finally, we compared the \emph{Robot Arm Repertoire}, with $10000$ archive size, and IsoLineDD, using the different dimensions that we mentioned. Once again, we noticed some more significant differences (\autoref{fig:appendix:dim:robotarm} f18: $p < 1\mathrm{e}-21$ for all pairs, Mann-Whitney U test). As in the previous experiment, we have also noticed that the features after $1M$ evaluations, are significantly different than after $10k$ evaluations (\autoref{fig:appendix:dim:robotarm} f37: 16 dimensions: $10k$ with $1M$ $p < 1\mathrm{e}-10$, Mann-Whitney U test), and we have encountered features that do not show significant change (\autoref{fig:appendix:dim:robotarm} f9: 16 dimensions: $10k$ with $1M$ $p = 0.10$, Mann-Whitney U test).

% We can discuss and note some observations. 

% Add the conlusion at the end.

\section{Discussion and Conclusion}
We would like to emphasise, that some features (ELA convexity, local search, and curvature) require additional evaluations, which, in practice, could be very costly, as they require more computational resources. In addition, features like ELA curvature, meta-model, and cell mapping, are time intensive, due to calculations of gradients and Hessian matrices. Features that provide such issues could be either discarded, or optimised for QD. Future works, could introduce alternative features, more suitable for QD, that do not have high computational requirements. 
%Moreover, other operators or QD algorithms could be evaluated using ELA features, to provide a better understanding on how each of these algorithms behaves. 
%Lastly, we also want to emphasise, that more domains should be examined, using different operators, understand how the behave, which can guide us at which point an operator is most useful.
Lastly, by expanding our methodology to cover more domains and QD algorithms (or operators) could shed light on when each algorithm is most effective.

We have shown that by using \emph{Exploratory Landscape Analysis} for \emph{Quality Diversity} problems, we can see some significant changes on the characterisations on the Elite Hypervolume, when using different operators, behaviour functions, archive sizes, or genotypic dimensions. This proves that our hypothesis that QD affects the low-level ELA features in a way different that LHS does, is indeed correct, which shows that their usage can only be beneficial for the field of QD optimisation.

\bibliographystyle{ACM-Reference-Format}
\bibliography{refs}

% \pagebreak
\pagebreak
\appendix
\section{Features}
\begin{table}[h]
    \centering
    \begin{tabular}{ll}
    \toprule
    Feature & Code \\
    \midrule
    ela\_conv.conv\_prob & f1 \\
    ela\_conv.lin\_dev\_abs & f2 \\
    ela\_conv.lin\_dev\_orig & f3 \\
    ela\_conv.lin\_prob & f4 \\
    ela\_distr.kurtosis & f5 \\
    ela\_distr.number\_of\_peaks & f6 \\
    ela\_distr.skewness & f7 \\
    ela\_level.lda\_qda\_10 & f8 \\
    ela\_level.lda\_qda\_25 & f9 \\
    ela\_level.lda\_qda\_50 & f10 \\
    ela\_level.mmce\_lda\_10 & f11 \\
    ela\_level.mmce\_lda\_25 & f12 \\
    ela\_level.mmce\_lda\_50 & f13 \\
    ela\_level.mmce\_qda\_10 & f14 \\
    ela\_level.mmce\_qda\_25 & f15 \\
    ela\_level.mmce\_qda\_50 & f16 \\
    ela\_local.basin\_sizes.avg\_best & f17 \\
    ela\_local.basin\_sizes.avg\_non\_best & f18 \\
    ela\_local.basin\_sizes.avg\_worst & f19 \\
    ela\_local.best2mean\_contr.orig & f20 \\
    ela\_local.best2mean\_contr.ratio & f21 \\
    ela\_local.n\_loc\_opt.abs & f22 \\
    ela\_local.n\_loc\_opt.rel & f23 \\
    ela\_meta.lin\_simple.adj\_r2 & f24 \\
    ela\_meta.lin\_simple.coef.max & f25 \\
    ela\_meta.lin\_simple.coef.max\_by\_min & f26 \\
    ela\_meta.lin\_simple.coef.min & f27 \\
    ela\_meta.lin\_simple.intercept & f28 \\
    ela\_meta.lin\_w\_interact.adj\_r2 & f29 \\
    ela\_meta.quad\_simple.adj\_r2 & f30 \\
    ela\_meta.quad\_simple.cond & f31 \\
    ela\_meta.quad\_w\_interact.adj\_r2 & f32 \\
    nbc.dist\_ratio.coeff\_var & f33 \\
    nbc.nb\_fitness.cor & f34 \\
    nbc.nn\_nb.cor & f35 \\
    nbc.nn\_nb.mean\_ratio & f36 \\
    nbc.nn\_nb.sd\_ratio & f37 \\
    \bottomrule
    \end{tabular}
    \caption{Mappings of the features returned by pflacco.}
    \label{table:features}
\end{table}

% Add the legend to the side (open space).
\begin{landscape}
\section{Figures}\label{appendix:figures}
    \begin{figure}[ht]
        \centering
        \includegraphics[scale=0.725]{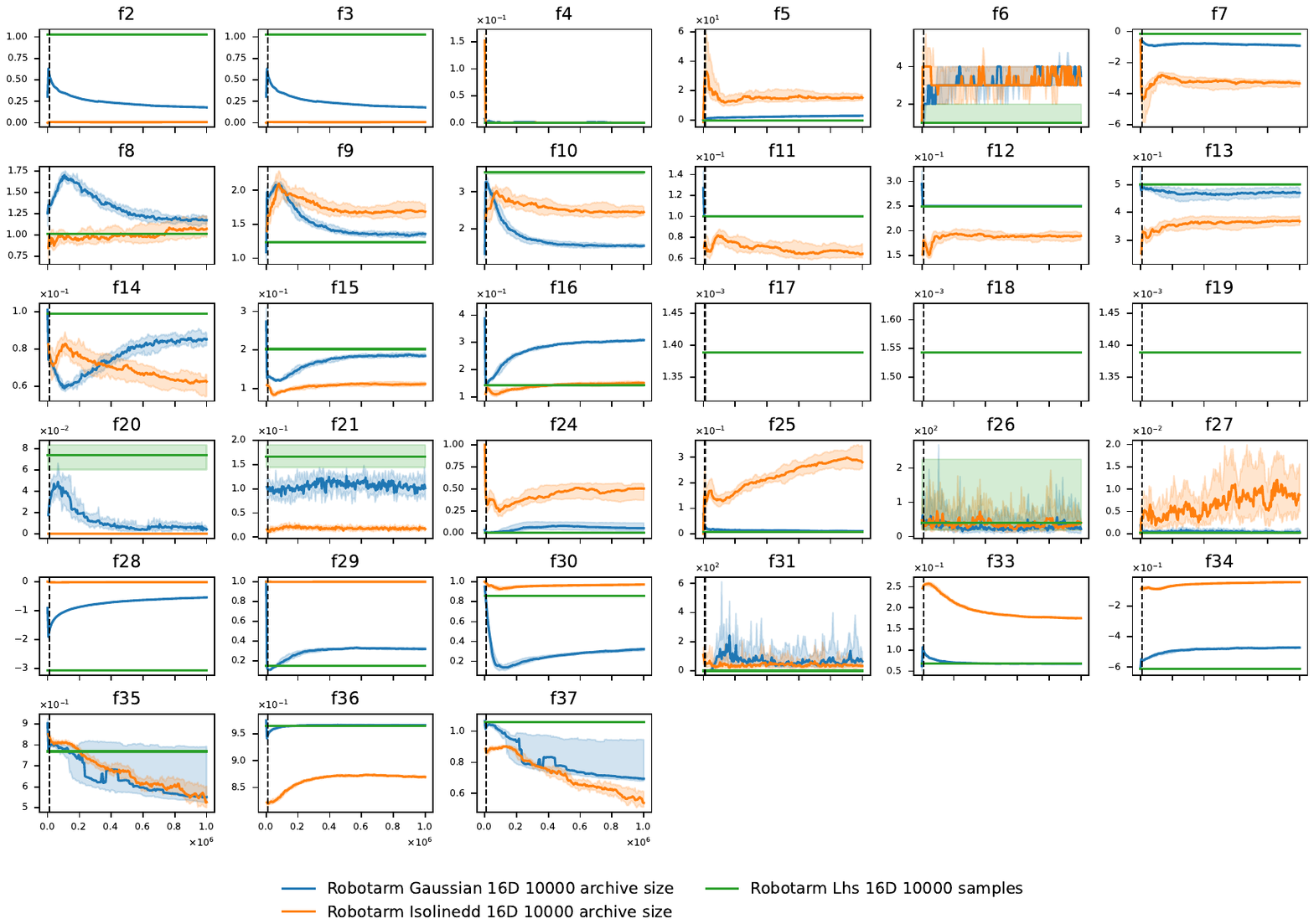}
        \caption{Gaussian mutation operator and IsoLineDD in comparison with LHS with $10000$ archive size on the \emph{Robot Arm Repertoire} domain with $16$ dimensions. The plots show the median and the interquartile range ($Q_1$ and $Q_3$) over $30$ experimental runs, with mostly notable differences. The x-axis, represents the total number of evaluations, whilst the vertical dashed line marks the $10000$ evaluations, matching the $10000$ evaluations of LHS.}
        \Description{Gaussian multivariate operator and IsoLineDD in comparison with LHS with $10000$ archive size on the \emph{Robot Arm Repertoire} domain with $16$ dimensions. The plots show the median and the interquartile range ($Q_1$ and $Q_3$) over $30$ experimental runs, with mostly notable differences. The x-axis, represents the total number of evaluations, whilst the vertical dashed line marks the $10000$ evaluations, matching the $10000$ evaluations of LHS.}
        \label{fig:appendix:operator:dim16}
    \end{figure}
\end{landscape}

% \section{Dimensionality Figures}
% \begin{figure*}[h]
%     \centering
%     \includegraphics[width=0.9\textwidth, height=0.9\textheight]{figures/robotarm_isolinedd_dim-all_samples10000.pdf}
%     \label{fig:enter-label}
%     \caption{CVT-MAP-Elites + IsoLineDD with 10000 archive cells on Robot Arm domain with different dimensions.}
%     \Description{Figure description}
% \end{figure*}

% \begin{figure*}[h]
%     \centering
%     \includegraphics[width=0.9\textwidth, height=0.9\textheight]{figures/robotarm_gaussian_dim-all_samples10000.pdf}
%     \label{fig:enter-label}
%     \caption{CVT-MAP-Elites + ISO (Gaussian) with 10000 archive cells on Robot Arm domain with different dimensions.}
%     \Description{Figure description}
% \end{figure*}

\end{document}